\documentclass[
]{ceurart}

\usepackage{booktabs}
\usepackage{amsmath}
\usepackage{graphicx}
\usepackage{enumitem}
\usepackage{microtype}
\usepackage{tikz}
\usetikzlibrary{arrows.meta,positioning,shapes.geometric}
\usepackage{float}

\begin{document}

\copyrightyear{2026}
\copyrightclause{Copyright for this paper by its authors.
  Use permitted under Creative Commons License Attribution 4.0
  International (CC BY 4.0).}

\conference{CSEDM'26: 10th Workshop on Computer Science Educational Data Mining,
  co-located with the 18th International Conference on Educational Data Mining
  (EDM 2026), 2026}

\title{Detecting Knowledge Gaps from Conversational AI Interactions
Using Curriculum Prerequisite Graphs}

\author[1]{Youssef Medhat}[%
orcid=0009-0002-3343-0149,
email=ymedhat3@gatech.edu,
]
\cormark[1]

\author[1]{Junsoo Park}[%
orcid=0009-0003-4450-0103,
email=jpark3232@gatech.edu,
]

\author[1]{Ploy Thajchayapong}[%
orcid=0009-0000-5993-1094,
email=ploy@gatech.edu,
]

\author[1]{Ashok K. Goel}[%
orcid=0000-0003-4043-0614,
email=ashok.goel@cc.gatech.edu,
]

\address[1]{Georgia Institute of Technology, Atlanta, GA, USA}

\cortext[1]{Corresponding author.}

\begin{abstract}
Large online courses generate thousands of student questions directed at
conversational AI teaching assistants, yet these interaction logs remain
largely untapped as diagnostic signals.
We present a pipeline that maps student questions from a conversational AI
teaching assistant to curriculum topics using a few-shot text classifier,
grounded in a GPT-4-extracted prerequisite knowledge graph of course concepts.
Evaluated on 1,340 question events from 164 students in a graduate-level AI course,
our classifier achieves 80.0\% accuracy across 43 labels (42 curriculum topics plus an ``unknown'' abstention class).
Topic-level question volume correlates significantly with student
self-reported difficulty from an independent mid-semester survey
($\rho = 0.491$, $p = 0.008$, $n = 28$ topics),
providing convergent evidence that the classified question stream
reflects genuine topic difficulty.
These results demonstrate that conversational AI interaction logs, mapped
onto curriculum structure, carry actionable signals about topic-level
knowledge gaps and provide instructors with a curriculum-grounded view
of which topics warrant attention.
\end{abstract}

\begin{keywords}
  knowledge gap detection \sep
  conversational AI teaching assistants \sep
  curriculum knowledge graph \sep
  prerequisite structure \sep
  few-shot text classification \sep
  educational data mining \sep
  learning analytics
\end{keywords}

\maketitle

\section{Introduction}

Conversational AI teaching assistants have been deployed in large-scale online
courses to answer student questions around the clock~\cite{goel2018jillwatson}.
Each submitted question is a behavioral trace: it reveals which topic a student
is thinking about, when, and, indirectly, whether they have understood prior
material~\cite{maiti2025criticalquestions}.
Despite the scale of these interaction logs, relatively little prior work
has analyzed them for individualized diagnostic purposes beyond aggregate
dashboards~\cite{fu2025mininggold,cabral2025lads}.

Existing approaches to knowledge gap detection rely on structured item
responses (correct or incorrect answers to assessments), which require
carefully designed item banks aligned to a skill taxonomy~\cite{corbett1994bkt,piech2015deep}.
Free-form conversational questions are harder to analyze: they must first be
mapped to curriculum topics, and the diagnostic signal is implicit.
A student asking many questions about a topic, or asking about it weeks after
it was taught, is exhibiting a behavioral signature of difficulty, but
extracting this signal requires both a text classifier and a curriculum model.

This paper presents a pipeline that bridges these two requirements.
We use a GPT-4-extracted prerequisite knowledge graph of course topics and
a few-shot text classifier (FastFit~\cite{yehudai2024fastfit}) to transform
raw student questions from a conversational AI teaching assistant into
curriculum-grounded topic-level signals.
The design choice of treating \emph{topics} (not individual questions)
as the analytic unit is motivated by how student struggle actually
manifests in coursework.
Students frequently show difficulty with a particular lesson or topic,
but the underlying cause is often not local to that topic: it lies in an
upstream prerequisite that was not fully consolidated.
Treating questions as topic-level signals, and overlaying them on a
prerequisite knowledge graph, is what makes these structurally
grounded knowledge gaps visible at all.
Topic-level aggregation additionally matches the granularity at which
an instructor can act and is empirically more robust to classifier
noise than per-question correctness assessment would be.
This work-in-progress is organized around a single research question:

\noindent\textbf{RQ.} \emph{Can naturally occurring student questions to a
conversational AI teaching assistant, mapped onto a curriculum
prerequisite knowledge graph, surface topic-level knowledge gaps that
match an independent measure of topic difficulty?}

\noindent We decompose the RQ into two hypotheses:
\begin{itemize}[leftmargin=1.2em,itemsep=1pt]
  \item \textbf{H1} (Classification): A few-shot classifier maps questions
        to curriculum topics with sufficient accuracy for downstream
        topic-level analysis.
  \item \textbf{H2} (Survey Alignment): Topic-level question volume from
        the classified stream correlates with self-reported difficulty
        from an independent survey.
\end{itemize}

\noindent Support for H1 establishes that the upstream measurement step
is reliable, and support for H2 then constitutes evidence that the
resulting topic-level signal reflects genuine course-level knowledge
gaps: those topics on which the cohort, in aggregate, struggles.

\begin{figure}[!ht]
\centering
\begin{tikzpicture}[
  stepbox/.style={
    draw=black!65, rounded corners=5pt, fill=blue!5,
    minimum width=0.93\linewidth, text width=0.91\linewidth,
    align=left, inner sep=8pt, font=\small
  },
  arrow/.style={-{Stealth[length=6pt,width=5pt]}, thick, black!55},
  node distance=6pt
]
\node[stepbox] (s12) {%
  \textbf{Steps 1--2 \enspace Knowledge Graph \& Labeling}%
  \hfill\textit{[GPT-4]}\\[2pt]
  Lesson PDFs $\to$ prerequisite graph: \textbf{54 topics, 47 edges.}\enspace
  GPT-4 labels 1,046 student questions.
};
\node[stepbox, below=of s12] (s34) {%
  \textbf{Steps 3--4 \enspace Classifier Training \& Evaluation}%
  \hfill\textit{[H1]}\\[2pt]
  FastFit batch-contrastive fine-tuning, 43 classes.
  Gold-set: \textbf{80.0\% accuracy, macro F1\,=\,0.669.}
};
\node[stepbox, below=of s34] (s5) {%
  \textbf{Step 5 \enspace Survey Validation}%
  \hfill\textit{[H2]}\\[2pt]
  Question volume vs.\ self-reported difficulty (280 respondents, 28 topics):
  \textbf{$\rho=0.491$,} $p=0.008$.
};
\draw[arrow] (s12.south) -- (s34.north);
\draw[arrow] (s34.south) -- (s5.north);
\end{tikzpicture}
\caption{Pipeline overview. H-labels indicate where each hypothesis is evaluated.}
\label{fig:pipeline}
\end{figure}
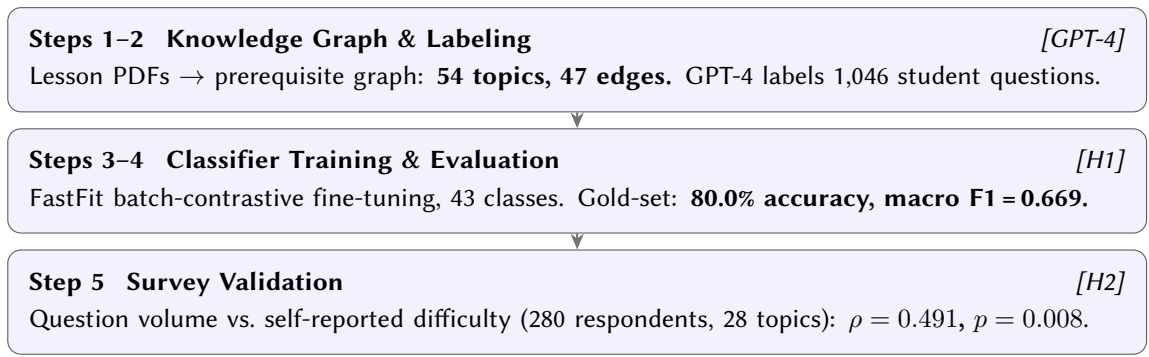

\section{Related Work}

\textbf{Knowledge tracing.}
Bayesian Knowledge Tracing~\cite{corbett1994bkt} and Deep Knowledge
Tracing~\cite{piech2015deep} estimate per-skill mastery from structured
item-response sequences.
Extensions incorporate prerequisite structure~\cite{chen2018prereq}.
These approaches require aligned item banks and cannot be applied to the
free-form conversational questions we study here.

\textbf{Prerequisite structure in curricula.}
Prerequisite relationships between topics have been extracted from course
syllabi~\cite{pan2017prereq}, textbooks~\cite{liang2018prereq}, and
knowledge graphs.
Prior work uses these structures for curriculum design and concept
recommendation, but not for gap detection from conversational logs.

\textbf{Conversational AI in CS education.}
Our work differs from prior conversational-AI deployments in CS education
in two specific respects: (i)~we pair the chat-log signal with an
LLM-extracted curriculum prerequisite knowledge graph, and (ii)~we
validate the resulting topic-level signal against an independent
self-reported difficulty survey rather than against engagement or
retention proxies.
Prior work in this space has largely deployed intelligent tutoring
systems and conversational agents in programming courses and STEM
contexts~\cite{roll2011metacognition}.
Recent deployments of conversational AI teaching assistants in large
online courses field thousands of student queries per semester, and
recent work has begun to analyze these chat logs for automated
knowledge gap detection~\cite{fu2025mininggold} and for
learning-analytics dashboarding~\cite{cabral2025lads}.
Complementary lines of work in our group examine the representational
and behavioral aspects of these deployments: learner representations
derived from chat-log interactions and their evaluation for downstream
differentiation~\cite{park2026representations}, the effect of
memory-based versus context-only conditioning on the personalization
behavior of LLM-based agents~\cite{park2026memory}, and a real-world
classroom assessment of bidirectional feedback in multi-tier AI
deployments~\cite{basu2025bidirectional}.

\textbf{Few-shot text classification.}
Sentence-level embedding models such as Sentence-BERT~\cite{reimers2019sentencebert}
have enabled accurate text classification from limited labeled data
by reusing semantic similarity between pre-trained embeddings.
SetFit~\cite{tunstall2022setfit} and FastFit~\cite{yehudai2024fastfit}
build on this by adapting the embedding space with contrastive objectives,
achieving competitive accuracy with as few as a single labeled example per class.
Our use of FastFit is motivated by this property: several curriculum topics
in our setting have highly imbalanced training coverage.

\textbf{LLM-assisted curriculum mining.}
Large language models have recently been used to extract structured
educational artifacts from unstructured course content, including
concept lists, definitions, and prerequisite relations~\cite{liang2018prereq}.
We use GPT-4 in this role to bootstrap both the knowledge graph
and an initial pool of weakly labeled questions; a human-annotated
gold set is used to evaluate the downstream classifier.

\section{Method}

\subsection{Knowledge Graph Construction}

\textbf{Role in the pipeline.}
The prerequisite knowledge graph serves two distinct roles in this
work-in-progress.
First, the \emph{topic nodes} provide the closed label space the rest
of the pipeline depends on: H1 measures classifier accuracy against a
gold set whose labels are drawn from this set, and H2 correlates
topic-level question volume with the survey's self-reported difficulty
by aligning topic names across the two instruments.
Both analyses require a fixed, named topic set; the knowledge-graph
extraction is what produces it.
Second, the \emph{directed prerequisite edges} encode the curriculum
structure that future per-student gap analyses can propagate over:
when a student is observed to have difficulty on a downstream topic,
the upstream prerequisites in the graph identify candidate root-cause
topics that merit review.
The current paper focuses on the topic-level signal (H1 and H2);
we treat the propagation of per-student gap signals along prerequisite
edges as the immediate next analytical step on this data, described
in Section~\ref{sec:future}.

A course instructor with domain expertise used GPT-4 to extract topics
and prerequisite relationships from 26 lesson PDFs of a graduate-level
AI course at a US R1 university, following a three-stage protocol.
\textbf{(i)~Topic extraction:} each PDF was processed individually and
GPT-4 was prompted to list the distinct teachable units introduced in
the lesson along with a one-sentence operational definition.
\textbf{(ii)~Topic boundary consolidation:} near-identical topics
across lessons were merged; topics the textbook treats as distinct
(e.g., Case-Based Reasoning vs.\ Analogical Reasoning) were kept
separate even when vocabulary overlapped.
\textbf{(iii)~Edge extraction:} for each candidate pair, GPT-4 was
given the two definitions and asked whether one is a direct
prerequisite for the other; ambiguous or bidirectional responses were
discarded.
The resulting graph contains \textbf{54 topic nodes} and
\textbf{47 directed prerequisite edges} (e.g.,
\textit{Semantic Networks} $\to$ \textit{Frames};
\textit{Case-Based Reasoning} $\to$ \textit{Analogical Reasoning};
\textit{Logic} $\to$ \textit{Explanation-Based Learning}).
The 26 lesson PDFs and the resulting node/edge counts are summarized
in Figure~\ref{fig:topic-structure}.

\textbf{Graph validation.}
Two pieces of evidence support treating the resulting graph as
defensible for the downstream analyses rather than as raw LLM output.
First, the edge-extraction stage \emph{discards} every GPT-4
prerequisite judgment that is bidirectional or ambiguous, so the
graph only contains the subset of relations the LLM was unambiguous
about.
Second, representative edges (\textit{Semantic Networks}
$\to$ \textit{Frames}; \textit{Case-Based Reasoning}
$\to$ \textit{Analogical Reasoning}) match the lesson ordering in the
course textbook and prior pedagogical work on the same curriculum.
We do not claim the resulting graph is the unique correct structure;
the protocol is a low-cost, instructor-auditable way to obtain a
topic--edge artifact that supports the downstream analyses without
the manual effort of writing the graph from scratch.

\begin{figure}[!ht]
\centering
\footnotesize
\setlength{\tabcolsep}{4pt}
\begin{tabular}{@{}lll@{}}
\toprule
\textbf{Lesson block} & \textbf{Representative topics} & \textbf{Sample prerequisite edges} \\
\midrule
Foundations & Foundations of AI; Cognitive Architectures
  & Cognitive Architectures $\to$ Frames \\
Knowledge rep.\ & Semantic Networks; Frames; Logic
  & Semantic Networks $\to$ Frames \\
Problem solving & Means-Ends Analysis; Planning; Sub-Goals
  & Planning $\to$ Sub-Goals \\
Learning & Generalization; Version Spaces; EBL
  & Logic $\to$ Explanation-Based Learning \\
Reasoning & CBR; Analogical; Commonsense
  & Case-Based $\to$ Analogical Reasoning \\
Agents/ethics & Properties of AI Agents; Ethics
  & Foundations of AI $\to$ AI Agents \\
\bottomrule
\end{tabular}
\caption{Representative slice of the 54-topic prerequisite knowledge
graph extracted by GPT-4 from 26 lesson PDFs. Each row groups topics
taught in the same lesson block with a sample prerequisite edge; the
full graph has 47 directed edges.}
\label{fig:topic-structure}
\end{figure}

\subsection{Question Classification}

\textbf{LLM-assisted labeling.}
Training labels for the 1,046-question pool were produced by GPT-4 under a
closed-label protocol: the prompt lists all 54 topics with their one-sentence
definitions from the knowledge-graph stage, followed by the student question,
and asks for the single best topic label or the literal string
\texttt{unknown} when no topic fits.
Because the label space is fixed and enumerated in the prompt, GPT-4 cannot
introduce new topics, and the downstream classifier inherits the same bounded
label space.

\textbf{Why not use the LLM directly as the classifier?}
GPT-4 produces the training labels in the previous step, so it is
natural to ask why we add a downstream classifier rather than calling
GPT-4 at inference time.
We retain the FastFit classifier for two reasons.
First, GPT-4 at inference time is non-deterministic and externally
versioned; FastFit gives reproducible predictions across runs.
Second, on the same 70-question gold set with the same 43-class label
space, GPT-4 used directly under the same closed-label prompt scores
$0.786$ accuracy / $0.654$ macro F1, slightly under FastFit's
$0.800 / 0.669$ (Table~\ref{tab:h1}).
The downstream classifier therefore matches the LLM's direct accuracy
on this task while removing per-query LLM dependence.

\textbf{Model.}
We trained a FastFit~\cite{yehudai2024fastfit} classifier on the 1,046
LLM-labeled student questions.
FastFit uses batch contrastive learning over sentence embeddings to achieve
high accuracy with limited per-class examples, a critical property given
that some topics have as few as one training sample.
The classifier assigns each question to one of the 54 topic classes or an
``unknown'' label when confidence is insufficient.

\textbf{Backbone and preprocessing.}
We use \texttt{all-mpnet-base-v2}~\cite{reimers2019sentencebert} as the
frozen sentence-embedding backbone.
Each raw Caliper event is parsed to extract the student-authored question
string; system-generated prompts and duplicate questions within a single
session are removed prior to classification.

\textbf{Training configuration.}
FastFit is fine-tuned with a batch size of 32, a learning rate of
$1{\times}10^{-5}$, 40 epochs, and 5 training repeats per class.
The token-level similarity head is enabled; the sentence-transformer
backbone weights are held frozen.
Training runs in under 20 minutes on a single GPU.

\textbf{Class coverage.}
Of the 54 topic nodes in the knowledge graph, 42 receive at least one
LLM-assigned training label; together with an ``unknown'' abstention
class for off-topic questions, this gives an effective classifier
output space of 43 labels.
The remaining 12 topics have no training questions and are not
predicted.
Per-class label counts among the 42 substantive topics are highly
skewed (median = 5, max = 89, min = 1), reflecting the long-tailed
nature of student interest; the abstention class itself accumulates
464 examples.
To counteract this imbalance during training we downsample the
abstention class to 80 examples and augment each topic with fewer than
10 examples by three short paraphrases derived from its KG
description, yielding a balanced pool of 737 training items.

\subsection{Survey Validation and H2 Test}

For H2, we compare topic-level assistant question volume (total
questions per topic across all students) against self-reported
difficulty from an independent mid-semester Qualtrics survey
($n = 280$ respondents).
Both are aggregate class-wide signals, making raw question volume the
appropriate assistant-side counterpart to the aggregate survey
measure.
We rank both by topic and compute Spearman's rank correlation $\rho$
over the 28 topics present in both datasets.

\textbf{Matching survey topics to graph topics.}
The mid-semester survey item asks respondents to rank which topics
they found most difficult from a fixed list authored by the course
instructor in collaboration with the teaching team, prior to and
independently of the knowledge-graph extraction.
We construct the matched set of 28 topics by manually aligning each
survey topic to its corresponding node in the 54-topic graph.
Matching is straightforward where the survey topic name and graph
topic name agree verbatim (e.g., \textit{Analogical Reasoning},
\textit{Planning}); where wording differs but the underlying concept
is the same (e.g., the survey's \textit{Generate \& Test} maps to the
graph's \textit{Generate-and-Test}), we record the alignment in a
matching table that is shipped with the analysis code.
Survey topics with no clear graph counterpart (8 topics, mostly
late-semester project or administrative items) and graph topics with
no corresponding survey entry (18 topics, mostly low-traffic
specialized concepts) are excluded from the comparison.
The final matched set of 28 topics is fixed before the correlation
is computed.

\section{Evaluation}

\subsection{Dataset}

We analyze \textbf{1,340 question events} from \textbf{164 students} in a
graduate-level AI course offered online by a US R1 university.
A \emph{question event} is a single Caliper \texttt{MessageEvent} record
from a conversational AI teaching assistant in which the student actor
authored a non-empty question; system-generated prompts, internal
acknowledgements, and non-question utterances are excluded.
The 1,340 raw question events are deduplicated and filtered to
1,046 unique student-authored questions, which form the pool used for
downstream LLM annotation and classifier training.
Caliper events are ingested through the
A4L data architecture~\cite{goel2025a4l,thajchayapong2025a4levolution}
and processed by the configurable analytics
pipeline of \cite{santana2025pipeline}, which provides the per-cohort
event extraction and deduplication described below.
The mid-semester survey has \textbf{280 respondents} who self-reported
which topics they found most difficult.

\textbf{Data handling.}
All analyses were conducted on pseudonymized logs under an institutionally
approved research protocol.
Direct identifiers (name, email, student ID) were removed at ingestion and
replaced with opaque per-student hashes before any processing.
No raw question text is released with this paper; examples quoted in the
error analysis are paraphrased to remove content that could identify
individual students or cohorts.

\subsection{H1: Classification Accuracy}

The classifier was evaluated on an initial human-annotated gold set of 70
questions, labeled by a course teaching assistant using the course textbook
as ground truth.
The set is small relative to the 43 target classes; expanding it is a
priority for ongoing work.

\begin{table}[h]
\centering
\caption{Classifier evaluation on the 70-question gold set
(43-class label space). FastFit is our pipeline; GPT-4 direct is the
same closed-label prompt used to generate the training set, applied at
inference time as a baseline.}
\label{tab:h1}
\begin{tabular}{@{}lcc@{}}
\toprule
\textbf{Method} & \textbf{Accuracy} & \textbf{Macro F1} \\
\midrule
GPT-4 direct (closed-label prompt) & 0.786 & 0.654 \\
FastFit (ours)                     & \textbf{0.800} & \textbf{0.669} \\
\midrule
\multicolumn{3}{l}{\footnotesize FastFit weighted F1: 0.856.} \\
\bottomrule
\end{tabular}
\end{table}

\textbf{Error analysis.}
The 14 misclassified gold items cluster into three interpretable
patterns:
\emph{(i) Surface-keyword traps} (7/14), where a strongly cued word
pulls the prediction toward an unrelated topic (e.g., a question about
Schank's primitive actions classified as \textit{Classification}
rather than \textit{Commonsense Reasoning});
\emph{(ii) adjacent reasoning topics} (5/14), mostly gold-label
\textit{Analogical Reasoning} questions classified into linked
problem-solving topics (\textit{Problem Reduction},
\textit{Knowledge Representation}) whose textbook vocabulary
(``similarity'', ``mapping'', ``structural'') overlaps;
and \emph{(iii) vague questions and abstention misfires} (2/14), where
the question is too fragmentary to disambiguate or the abstention
class fires inappropriately.
Pattern (i) reflects a classifier limitation that scaling the training
pool would address; pattern (ii) reflects topic boundaries that a
multi-label extension could resolve.
The gap between macro F1 (0.669) and weighted F1 (0.856) is driven by
the long tail of singleton-support classes in the gold set, the
expected behavior of a few-shot classifier on a heavily imbalanced
class set.

\textbf{Per-topic performance.}
The \texttt{unknown} abstention class dominates support
(22/70 gold items) and is well classified ($F1 = 0.93$), confirming
that the model reliably filters off-topic chat (course logistics,
trolling, out-of-curriculum questions).
Among substantive topics, the higher-support classes (Analogical
Reasoning at $F1 = 0.55$ and Commonsense Reasoning at $F1 = 0.67$)
are where most misclassifications concentrate, since these topics
share vocabulary with adjacent reasoning topics in the curriculum and
account for most of the macro-F1 gap.

\textbf{H1 is supported}: 80\% accuracy appears sufficient for preliminary
aggregate topic-level analysis in this setting, and the error structure is
concentrated in predictable, interpretable failure modes.

\subsection{H2: Survey Alignment}

\begin{table}[h]
\centering
\caption{Survey validation: assistant question volume vs.\ self-reported difficulty.}
\label{tab:h2}
\begin{tabular}{@{}lc@{}}
\toprule
\textbf{Metric} & \textbf{Value} \\
\midrule
Spearman $\rho$ & 0.491 \\
$p$-value & 0.008 \\
Matched topics & 28 \\
Significant ($\alpha = 0.05$) & \textbf{Yes} \\
\bottomrule
\end{tabular}
\end{table}

Topic-level assistant question volume tracks self-reported difficulty significantly
across 28 matched topics.
Analogical Reasoning generates the most assistant questions (89 events) and is the
fifth most-reported survey difficulty topic (19.3\% of respondents).
Generalization (33 events) and Constraint Propagation (28 events) also rank
highly in both measures.

\textbf{Per-topic rank agreement.}
Grouping the 28 matched topics by absolute rank difference $|\Delta|$
between assistant volume rank and survey difficulty rank: 14 of 28
fall in strong agreement ($|\Delta| \leq 3$), 9 in moderate
disagreement ($4 \leq |\Delta| \leq 10$), and 5 are outliers
($|\Delta| > 10$).
The outliers split into two interpretable subpatterns: late-semester
meta-level topics that students rate hard but rarely query, and
concept-introduction topics that generate terminology questions out of
proportion to their conceptual difficulty.

\textbf{H2 is supported}: assistant question volume is a statistically significant
surrogate for aggregate topic difficulty.

\section{Discussion and Limitations}
\label{sec:limitations}

\textbf{Practical implication.}
The pipeline converts passively logged assistant interactions into
topic-level difficulty signals without additional student burden.
Instructors are shown which topics attract the most questions and can
verify the ranking against independent survey evidence.

\textbf{Relevance to teachers and the AIED setting.}
The student-side input is naturally occurring chat data, but the
evaluation is grounded in instructor-authored curriculum artifacts (the
prerequisite knowledge graph and the gold set): the model is trained
on student behavior but assessed against curriculum representations the
instructor signed off on, making the resulting signal actionable in an
AIED workflow without requiring an item bank or a Bayesian skill
model.
The primary use we envision is a cohort-wide instructor dashboard;
the same signal can be passed to the conversational assistant
\emph{as suggestions} for prerequisite review when a student asks
about a downstream topic with an unresolved prerequisite, or surfaced
through the LMS as student-specific recommendations.
A related line of work in our group surfaces isolated learners through
outcome-independent mediation of feedback between teachers and
students using AI~\cite{park2026isolated}, framing the teacher--AI
loop as the unit of analysis rather than the student--AI dyad alone.

\textbf{Why topic classification rather than per-question solving.}
Per-question correctness assessment in free-form student questions is
a multi-hop reasoning problem that current LLMs handle unreliably.
Aggregating classified questions over a student's history sidesteps
this: the diagnostic signal is the \emph{distribution of help-seeking
across topics} rather than the correctness of any individual question,
which is empirically more robust and matches what an instructor can
realistically act on.

\textbf{Limitations.}
Three caveats bound the present claims.
(i)~The 20\% classification error rate introduces label noise into the
topic-volume counts, concentrated in topics with overlapping vocabulary
(e.g., Analogical Reasoning vs.\ Knowledge Representation); this likely
attenuates the measured correlation.
(ii)~Evaluation is on one graduate-level AI course at a US R1 university;
replication across courses and disciplines is needed.
(iii)~The H2 result is aggregate at the topic level and does not by itself
identify which individual students are struggling on which topic; we
return to this point in the Future directions paragraph below.

\textbf{Future directions.}
\label{sec:future}
The topic-level volume signal reported here points naturally toward
three complementary extensions on the same data.
First, the \emph{per-student gap signal}: students whose question
distribution over topics diverges anomalously from the cohort can be
flagged for targeted support, in contrast to the aggregate ranking
this paper reports.
Second, \emph{prerequisite-edge gap propagation}: the directed edges
in the knowledge graph become diagnostically useful once per-student
gaps are available, because an observed gap on a downstream topic can
be traced back along its prerequisite chain to candidate upstream
root-cause topics that the student has not consolidated.
This is the analytical step the graph edges are designed for; we are
preparing a companion analysis on the same cohort.
Third, a closed-loop \emph{intervention study} on the resulting
instructor-facing dashboard: surfacing per-student gap chains
alongside classifier confidence would allow a controlled test of
whether curriculum-grounded remediation prompts measurably improve
outcomes relative to a generic-help control.

\section{Conclusion}

We presented a work-in-progress pipeline that maps conversational AI
student questions onto a GPT-4-extracted prerequisite knowledge graph
via a few-shot classifier, and tested whether such logs surface
topic-level knowledge gaps matching an independent difficulty measure.
Both hypotheses are supported: 80.0\% classifier accuracy on a
43-class few-shot task (H1), and significant correlation between
assistant question volume and self-reported difficulty (H2:
$\rho = 0.491$, $p = 0.008$).
At the \emph{course level}, the topics that draw the most assistant
questions are, with statistical reliability, the same topics students
report as most difficult---evidence that conversational AI interaction
logs, mapped onto curriculum structure, surface course-level knowledge
gaps without requiring an item bank or per-question correctness
assessment.

\begin{acknowledgments}
We thank members of the A4L team in Georgia Tech's Design Intelligence
Lab (DILab; \texttt{dilab.gatech.edu}) for their contributions to this
research.
This research is supported by a US National Science Foundation grant
(Grant No.\ 2247790) to the National AI Institute for Adult Learning
and Online Education (AI-ALOE; \texttt{aialoe.org}).
\end{acknowledgments}

\section*{Declaration on Generative AI}

During the preparation of this work, the author(s) used Claude in
order to: Draft content, Paraphrase and reword.
After using this tool/service, the author(s) reviewed and edited the
content as needed and take(s) full responsibility for the publication's
content.


\end{document}